\documentclass[journal,twoside,web]{ieeecolor}
\usepackage{tmi}
\usepackage{cite}
\usepackage{amsmath,amssymb,amsfonts}
\usepackage{bbm, dsfont}
\usepackage{algorithmic}
\usepackage{graphicx}
\usepackage{textcomp}
\usepackage{multirow}
\usepackage{booktabs}
\usepackage[table,xcdraw]{xcolor}
\usepackage{arydshln}
\usepackage{url}
\usepackage{afterpage}
\usepackage{makecell}
\usepackage{hyperref}
\hypersetup{
    colorlinks=false,
    pdfborder={0 0 0}
}

\def\BibTeX{{\rm B\kern-.05em{\sc i\kern-.025em b}\kern-.08em
    T\kern-.1667em\lower.7ex\hbox{E}\kern-.125emX}}
\begin{document}
\title{PathScale-R1: Cross-scale Reasoning for Pathological Image Analysis}
\author{
Chi Phan,
Tianyi Zhang, 
Yufeng Wu,
Qiaochu Xue,
Jiajie Zhang,
Linghan Cai,
Zeyu Liu,\\
Sudong Wang,
Yueming Jin,~\IEEEmembership{Member, IEEE},
and Dan Hu
\thanks{This work was supported by the Ministry of Education Tier 1 grant, Singapore (24-1250-P0001), and the Ministry of Education Tier 2 grant, Singapore (T2EP20224-0028). This work was powered by the UnPuzzle \& PuzzleCloud Platform (https://puzzlelogic.com/unpuzzle) and supported by PuzzleLogic Pte Ltd, Singapore.}
\thanks{Chi Phan and Tianyi Zhang contributed equally to this work. Corresponding Author: Sudong Wang (e-mail: sudongwang@puzzlelogic.com), Yueming Jin (e-mail: ymjin@nus.edu.sg) and Dan Hu (e-mail: hudan@fjmu.edu.cn).}
\thanks{Chi Phan, Tianyi Zhang, and Yueming Jin are with Department of Electrical and Computer Engineering, National University of Singapore, Singapore 117417 (e-mails: \{chiphan, zhangtianyi\}@u.nus.edu; ymjin@nus.edu.sg).}
\thanks{Yufeng Wu, Linghan Cai, Zeyu Liu, and Sudong Wang are with PuzzleLogic Pte Ltd, Singapore 229594 (e-mails: \{yufengwu, linghancai, zeyuliu, sudongwang\}@puzzlelogic.com).}
\thanks{Qiaochu Xue and Yueming Jin are with Department of Biomedical Engineering, National University of Singapore, Singapore 117417, Singapore (e-mails: e1352520@u.nus.edu; ymjin@nus.edu.sg).}
\thanks{Jiajie Zhang and Dan Hu are with Department of Pathology, Fujian Medical University Cancer Hospital \& Fujian Cancer Hospital, Fuzhou, China (e-mails: naili816@126.com; hudan@fjmu.edu.cn).}
}

\maketitle

\newcommand{\datasetname}[1]{\text{#1}}
\newcommand{\modelname}[1]{\text{#1}}
\newcommand*{\dataset}{\datasetname{PathScale-VQA}}
\newcommand*{\model}{\modelname{PathScale-R1}}

\newcommand*{\mcq}{10{,}373} 
\newcommand*{\wsis}{373} 
\newcommand*{\triplets}{1{,}368} 
\definecolor{pinky}{HTML}{FFE2E1}
\definecolor{aliceblue}{rgb}{0.94, 0.97, 1.0}

\begin{abstract}
Pathological diagnosis is inherently multi-scale, requiring the integration of global tissue architecture at low magnification with cellular morphology at higher magnification. However, existing pathology benchmarks and vision-language models (VLMs) are still largely developed under single-scale settings, limiting their ability to learn clinically meaningful multi-magnification reasoning. Moreover, naively constructed visual question answering (VQA) tasks may be susceptible to text-only or superficial visual shortcuts, leading to unreliable assessments of visual understanding. To address these limitations, we introduce a benchmark and training framework for shortcut-resistant cross-scale pathology reasoning. We design an Adversarial Text-only Screening strategy for semantic reasoning questions and a Structure-controlled Distractor Sampling strategy for visual grounding questions, encouraging models to rely on cross-scale visual evidence. Based on this pipeline, we construct \dataset{}, a high-quality cross-scale pathology VQA benchmark with \mcq{} multiple-choice questions grounded in \triplets{} diagnostic paths across multiple magnification levels. Building on the semantic reasoning set, \model{} is optimized through Difficulty-driven Reasoning Distillation supervised fine-tuning followed by reinforcement learning with a Scale-aware Reasoning Structure reward, which encourages the use of evidence across magnifications. Extensive experiments demonstrate state-of-the-art performance of \model{} on cross-scale reasoning tasks and effective transfer to conventional single-scale pathology VQA. Our code is available at \href{https://github.com/iMVR-PL/PathScale-R1}{PathScale-R1}.
\end{abstract}

\begin{IEEEkeywords}
Vision-language model, Reinforcement Learning, Reasoning, Pathological Image Analysis.
\end{IEEEkeywords}

\section{Introduction}
\label{sec:introduction}

\IEEEPARstart{P}{athology} is the gold standard for cancer diagnosis, providing essential evidence for disease characterization, treatment planning, and prognosis~\cite {goldstandard2}. Due to the hierarchical organization and heterogeneity of tissue morphology, pathological interpretation requires intensive visual analysis and substantial domain expertise. Recent developments in large language models and multimodal learning have motivated pathology-focused Vision-Language Models (VLMs)~\cite{clover, quilt, pathor1, pathreasoner, pathlens, slidechat, wsillava}, which show promising capabilities in pathological understanding, visual question answering (VQA), and diagnostic support. Despite these substantial advancements, there is still a noticeable gap between the underlying design of current models and the practical workflow of pathological assessment.

Real-world pathological diagnosis is fundamentally a cross-scale reasoning process, in which pathologists systematically navigate across different magnifications to synthesize a final decision~\cite{patho1},~\cite{patho2},~\cite{patho3}. Slide review typically begins at low power to assess global tissue architecture and lesion distribution, proceeds to intermediate magnifications to examine localized tissue organization, and finally uses high magnification to verify cellular morphology and fine-grained diagnostic features~\cite{patho4, patho5}. These observations are highly interdependent, with macroscopic findings directing subsequent zoom-in decisions and microscopic evidence validating earlier structural impressions. Ultimately, a diagnosis is synthesized by merging this architectural, structural, and cellular evidence across scales. Therefore, a clinically meaningful pathology VLM should not only recognize findings at one specific scale, but also needs to connect cross-scale evidence to support a coherent diagnostic conclusion. 
\begin{figure*}[t!]
    \centering
    \includegraphics[width=\linewidth]{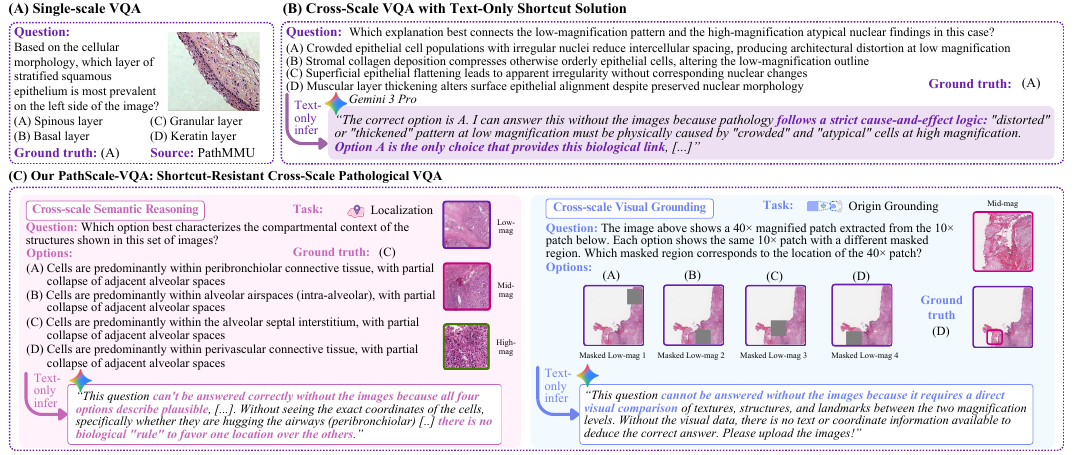}
    \caption{\textbf{Comparison of Pathology VQA designs.} (A) Single-scale VQA evaluates isolated local recognition.  (B) Naively constructed cross-scale VQA can still be solved from textual cues alone, revealing text-only shortcut risks. (C) Our \dataset{} uses expert-verified diagnostic paths to construct shortcut-resistant semantic reasoning and visual grounding tasks that require cross-scale visual evidence. (\textit{mag: magnification})
    }
    \label{fig:example}
\end{figure*}

However, existing pathology VLMs~\cite{clover, quilt, pathor1, pathreasoner, pathlens, slidechat, wsillava} are predominantly developed and evaluated in single-scale settings, which do not explicitly capture how diagnostically related evidence is connected across magnifications (Fig.~\ref{fig:example}). Patch-level VQA benchmarks such as PathVQA~\cite{pathvqa}, PathMMU~\cite{pathmmu}, and PathBench~\cite{pathbench-tmi} contain pathology images from diverse sources and magnification levels, but each question is typically associated with only a single ROI. They therefore assess recognition and reasoning within an isolated field of view, rather than whether a model can connect low-power tissue architecture, intermediate-scale tissue organization, and high-power cellular morphology from the same diagnostic trajectory. At the whole-slide image (WSI) level, benchmarks such as WSI-VQA~\cite{wsivqa}, SlideBench~\cite{slidechat}, and WSI-Bench~\cite{wsillava} evaluate slide-level understanding from gigapixel images. Existing WSI VLMs~\cite{wsillava, slidechat, pathreasoner} commonly make such inputs computationally tractable by aggregating patch embeddings at a specific magnification into a compact slide-level representation. Such representations support slide-level prediction, yet still provide a limited assessment of how findings across scales interact. Under these single-scale framings, models may learn scale-specific visual recognition, but receive little supervision for the hierarchical reasoning process that links low-power architectural context to high-power cellular evidence. Consequently, they may perform well on existing benchmarks while struggling in tasks that require cross-scale understanding, leading to a critical misalignment with clinical practice. Addressing this limitation requires moving beyond the current single-scale framing toward a cross-scale reasoning paradigm centered on recognizing and integrating evidence across magnifications.

Constructing cross-scale reasoning VQA tasks, however, encounters another critical challenge: the susceptibility to text-based shortcut solutions. In clinical pathology, prior knowledge and textual clinical context can guide interpretation, but assessments of tissue architecture, local organization, and cellular morphology are grounded in the visual evidence. Yet recent studies~\cite{textleak1,textleak2, textleak3} have shown that VLMs can achieve high benchmark performance by exploiting linguistic cues, weak distractors, or dataset-specific artifacts, rather than relying on the provided images. In medical VQA, such shortcuts may also yield fluent and seemingly image-grounded explanations despite limited visual dependence~\cite{mirage, pathbench-tmi}. This risk is particularly relevant in cross-scale pathology VQA, where magnification-specific terminology and stereotyped biomedical associations in the answer choices may reveal the correct response without requiring integration across scales (Fig.~\ref{fig:example}). Such leakage can create an ``illusion of visual understanding"~\cite{mirage}, making performance metrics unreliable for measuring model capabilities, which is particularly concerning in critical domains such as pathology. Therefore, constructing a reliable cross-scale reasoning setting requires not only collecting integrated multi-magnification images, but also actively suppressing shortcut solutions to enforce genuine reliance on the visual evidence.

\begin{figure*}[t!]
    \centering
    \includegraphics[width=1\linewidth]{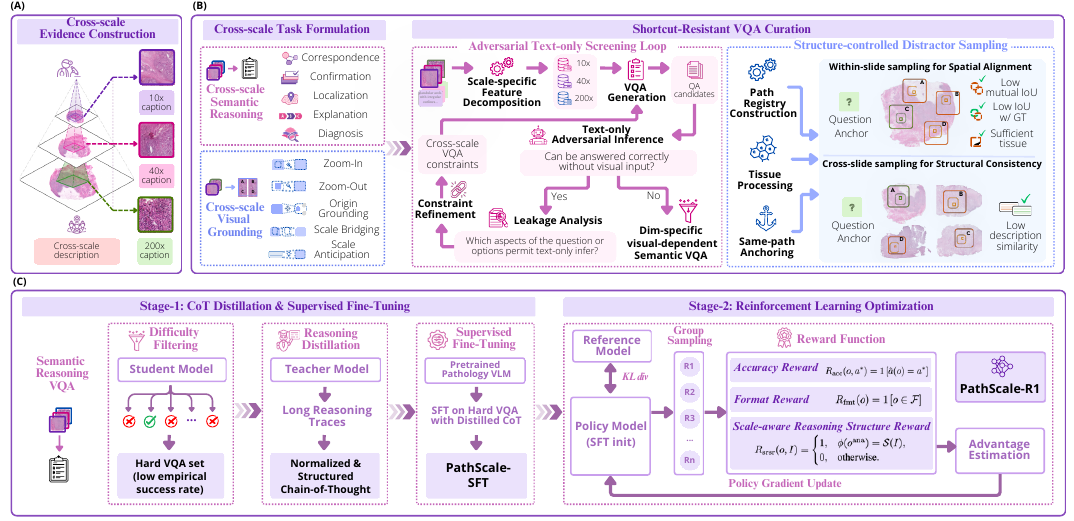}
    \caption{\textbf{Overview of the proposed cross-scale benchmark construction and model optimization framework.} (A) Expert-verified diagnostic paths link clinically relevant 10$\times$, 40$\times$, and 200$\times$ ROIs from the same WSI, providing scale-specific captions and cross-scale evidence anchors. (B) From these paths, we construct cross-scale semantic reasoning and visual grounding tasks, with adversarial text-only screening to reduce language shortcuts and structure-controlled distractor sampling to reduce superficial visual shortcuts. (C) \model{} is optimized by difficulty-driven reasoning distillation followed by reinforcement learning with accuracy, format, and scale-aware reasoning structure rewards.}
    \label{fig:main}
\end{figure*}

To address these challenges, we introduce a novel benchmark and training framework for shortcut-resistant cross-scale pathology reasoning (Fig.~\ref{fig:main}). \textit{\textbf{At the data level}}, we shift the basic unit of supervision and evaluation from an isolated ROI to a diagnostic path, defined as a pathologist-verified trajectory connecting clinically corresponding regions at 10\(\times\), 40\(\times\), and 200\(\times\) within the same WSI. To ensure that benchmark performance reflects genuine use of visual evidence, we formulate two complementary task categories, each paired with a targeted curation strategy: adversarial text-only screening for textual options, and structure-controlled distractor sampling for image options. \textit{\textbf{At the model level}}, we develop \model{}, a pathology VLM trained to integrate diagnostic evidence across scales through a two-stage framework. The first stage performs Difficulty-driven Reasoning Distillation, transferring structured cross-scale rationales on samples that are challenging for the base model. The second stage applies reinforcement learning with a novel Scale-aware Reasoning Structure reward, encouraging the model to produce coherent reasoning that incorporates evidence from multiple magnifications. Extensive experiments demonstrate that \model{} substantially improves cross-scale semantic reasoning while maintaining strong performance on conventional single-image pathology VQA benchmarks. An overview of the dataset and benchmark performance is provided in Fig.~\ref{fig:stats}.

The main contributions of this work are as follows:
\begin{itemize}
    \item We introduce a cross-scale formulation of pathological image analysis, in which the diagnostic path across magnifications serves as the basic unit of supervision and evaluation. Our curated cross-scale VQA benchmark, \textbf{PathScale-VQA}, comprises 10,373 questions grounded in 1,368 pathologist-verified diagnostic paths designed to reflect the practical workflow of pathological assessment.
    \item We develop a shortcut-resistant VQA curation pipeline that combines adversarial text-only screening with structure-controlled distractor sampling, reducing linguistic leakage and superficial visual cues for more reliable evaluation of cross-scale visual understanding.
    \item We propose \textbf{\model{}}, a two-stage optimization framework combining Difficulty-driven Reasoning Distillation with reinforcement learning guided by the novel Scale-aware Reasoning Structure Reward, enabling more effective integration of evidence across scales.
    \item Extensive experiments demonstrate that \model{} achieves state-of-the-art performance on the cross-scale benchmark and strong transferability to outperform the evaluated baselines on conventional single-scale pathology VQA. Our findings also further reveal that fine-grained cross-scale visual grounding is an important yet underdeveloped capability of current VLMs.
\end{itemize}

This work substantially extends our preliminary MICCAI 2026 version~\cite{scalereasoner} in four aspects. First, we expand the original benchmark from 4,685 to 10,373 questions, with substantially more annotated WSIs and diagnostic paths, thereby improving data coverage and clinical grounding. Second, we introduce five visual grounding tasks that reflect key operations in pathological assessment, together with structure-controlled distractors for a more rigorous evaluation. Third, we extend the optimization strategy to a two-stage framework consisting of difficulty-driven supervised fine-tuning with distilled rationales, followed by reinforcement learning guided by a novel Scale-aware Reasoning Structure Reward. Finally, we broaden the evaluation through additional model comparisons, ablation analyses, qualitative studies, and transfer experiments on conventional single-image pathology VQA. Together, these extensions establish a more comprehensive benchmark and training framework for advancing cross-scale reasoning in pathological image analysis.

\section{Related Work}
\subsection{Vision-Language Models for Pathology}
The rapid development of general-purpose VLMs, such as LLaVA~\cite{llava}, Qwen2.5-VL~\cite{qwen25vl}, Qwen3-VL~\cite{qwen3vl}, and InternVL3.5~\cite{internvl35}, has substantially improved multimodal understanding and visual reasoning. Medical VLMs, including LLaVA-Med~\cite{llavamed}, Lingshu~\cite{lingshu}, and HuatuoGPT~\cite {huatuogpt}, further adapt these capabilities to clinical images and biomedical question answering. Building on these developments, pathology-specific VLMs have emerged to support morphological interpretation, diagnostic support, and pathology-oriented reasoning. Patch-level models, such as Quilt-LLaVA~\cite{quilt}, CLOVER~\cite{clover}, PathAsst~\cite{pathasst}, PathLens~\cite{pathlens}, and Patho-R1~\cite{pathor1}, have demonstrated strong performance in pathology VQA, local feature recognition, and diagnostic reasoning. WSI-level models, including WSI-LLaVA~\cite{wsillava}, SlideChat~\cite{slidechat}, and PathReasoner-R1~\cite{pathreasoner}, extend this scope by aggregating information from multiple sampled regions to support slide-level understanding and reasoning over spatially distributed evidence. Despite these advances, explicit supervision for linking clinically corresponding evidence across magnification levels remains limited.
\subsection{Pathology Visual Question Answering Benchmarks}
Pathology VQA benchmarks have progressively expanded in scale, task diversity, and clinical relevance. Early patch-based datasets, such as PathVQA~\cite{pathvqa}, established the evaluation of natural-language question answering over histopathology images, while later benchmarks including Quilt-VQA~\cite{quilt} and PathMMU~\cite{pathmmu} introduced larger and more diverse question sets for assessing morphological recognition and diagnostic knowledge. WSI-level benchmarks, such as WSI-VQA~\cite{wsivqa}, SlideBench~\cite{slidechat}, and WSI-Bench~\cite{wsillava}, further extend evaluation to whole-slide understanding and slide-level question answering. Although these benchmarks cover a wide range of pathology tasks, dedicated evaluation of whether models can associate, verify, and integrate corresponding evidence across successive magnifications remains limited. Moreover, recent studies have shown that VQA benchmarks may be susceptible to linguistic priors, weak distractors, and annotation artifacts~\cite{textleak1,textleak2,mirage}, allowing models to exploit non-visual shortcuts. These gaps motivate \dataset{}, which evaluates cross-scale reasoning over clinically linked multi-magnification diagnostic paths with shortcut-resistant benchmark construction.

\section{Methodology}
\label{sec:methods}

\subsection{Cross-scale Diagnostic Path Construction}
To instantiate the principle of cross-scale evidence, we define the diagnostic path as the foundational evidence unit of our dataset. As illustrated in Fig.~\ref{fig:main}(A), a diagnostic path consists of a set of \{10$\times$,  40$\times$,  200$\times$\} ROIs from the same WSI. This structure reflects the clinical workflow of pathology diagnosis: low-power views provide architectural context, intermediate views reveal local tissue organization, and high-power views confirm cellular morphology. We construct these diagnostic paths from H\&E-stained WSIs obtained from The Cancer Genome Atlas (TCGA)~\cite{tcga}. Clinical experts are therefore involved at the earliest stage of our data construction, rather than serving only in post hoc validation as in prior benchmarks~\cite{pathmmu,wsillava,slidechat}. For each WSI, three junior pathologists independently inspect the slide to construct diagnostic paths. Each annotator first selects a diagnostically informative 10$\times$ ROI and then identifies clinically relevant subregions at 40$\times$ and 200$\times$, forming a progressive zoom-in trajectory across magnifications. These candidate paths are subsequently reviewed by a senior pathologist to ensure that each trajectory corresponds to a clinically meaningful diagnostic process. For each validated path, pathologists provide two levels of annotation: scale-specific captions describing the pathological features in each ROI and the rationale for each zoom-in decision. We then use GPT-5.2~\cite{gpt} to draft a cross-scale description from the ROI captions and zoom-in rationales. This description is refined and confirmed by pathologists to ensure that it explicitly connects low-power architecture, intermediate tissue organization, and high-power cellular morphology. The resulting expert-verified diagnostic paths and annotations serve as high-quality evidence anchors for subsequent VQA curation.

\subsection{Task Formulation \& Shortcut-resistant VQA curation }
Based on the expert-verified diagnostic paths, we construct two complementary VQA schemes: \textit{cross-scale semantic reasoning} and \textit{cross-scale visual grounding}, as illustrated in Fig.~\ref{fig:example}(C). The former evaluates whether models can integrate pathological evidence across magnifications, while the latter assesses visual correspondence between ROIs along a diagnostic trajectory. Because textual and image answer options are vulnerable to different shortcuts, we apply separate curation strategies to reduce text-only leakage and superficial visual cues, respectively.

\subsubsection{Cross-scale Semantic Reasoning}
In the semantic reasoning scheme, each sample is formulated as a multiple-choice VQA with textual answer options and a clinically linked set of ROIs from one diagnostic path. Each question is designed to require evidence from at least two magnifications, so the model must connect observations across scales instead of relying on single-view recognition. To cover the major reasoning steps involved in pathological diagnosis, we define five semantic task types. \textit{Correspondence} requires the model to determine whether findings observed at different magnifications represent the same pathological process. \textit{Confirmation} requires the model to determine whether higher-magnification evidence supports or contradicts a hypothesis formed at a lower magnification. \textit{Localization} requires the model to identify where diagnostically relevant evidence appears within the cross-scale image set. \textit{Explanation} requires the model to connect observations across magnifications to justify a conclusion. \textit{Diagnosis} requires the model to determine the final clinical interpretation supported by the combined cross-scale evidence. Together, these tasks provide a comprehensive evaluation of cross-scale diagnostic reasoning, from evidence association to final interpretation.
\begin{figure*}[t]
    \centering
    \includegraphics[width=1\textwidth]{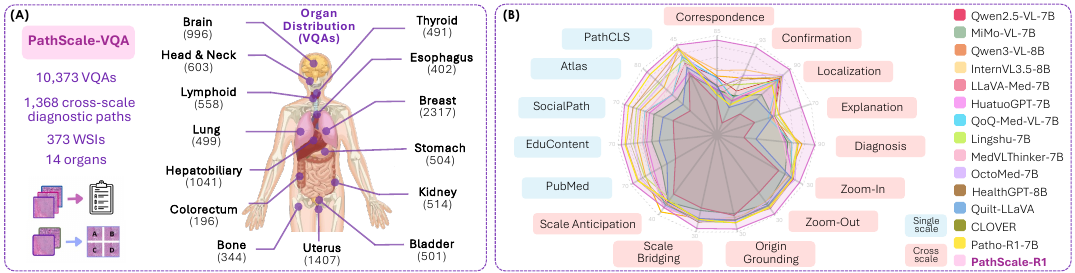}
    \caption{\textbf{Dataset statistics and benchmark performance.} (A) PathScale-VQA component statistics and organ distribution. (B) Task-wise performance of representative VLMs across single-scale and proposed cross-scale VQA benchmark.} 
    \label{fig:stats}
\end{figure*} 

\noindent\textbf{Adversarial Text-only Screening Loop.} 
Since semantic reasoning uses textual answer choices, the central curation challenge is to prevent the correct answer from being inferred from linguistic cues alone. To reduce this risk, we propose an adversarial generate-screen-revise loop for data curation (Fig.~\ref{fig:main}B). Specifically, we first decompose each diagnostic path into scale-specific feature sets, separating findings visible at 10$\times$, 40$\times$, and 200$\times$. Candidate questions are then generated from these decomposed feature sets under task-specific constraints tailored to each reasoning dimension. After generation, we subsequently remove all images and present only the question and textual options to strong closed-source LLMs, Gemini 3 Pro~\cite{gemini} and Qwen3-Max~\cite{qwen3max}. If either model predicts the correct answer, the sample is flagged as potentially solvable without visual evidence. For each flagged sample, we analyze the adversary's rationale to identify possible sources of leakage, including overly specific correct options, implausible distractors, imbalanced option granularity, answer-position bias, and diagnostic priors embedded in the question wording. We then revise the question, options, or generation constraints accordingly, and re-evaluate the revised sample under the same setting. This iterative process turns semantic VQA construction from one-pass generation into leakage-aware curation. Finally, the revised VQAs are reviewed by pathologists to verify that the designated answer is clinically correct, the required evidence is observable in the corresponding ROIs, and the question depends on findings from at least two magnifications. This process yields a semantic VQA set with reduced text-only solvability and stronger reliance on cross-scale visual evidence.

\subsubsection{Cross-scale Visual Grounding}
While semantic reasoning evaluates cross-scale interpretation through textual answer choices, visual grounding evaluates whether a model can recover the visual relationships that support such interpretation. We formulate this scheme as image-option MCQs derived from expert-verified diagnostic paths. Each sample contains an anchor and multiple visual alternatives, with the correct option defined by the corresponding cross-scale trajectory. By replacing textual options with visual alternatives, this scheme reduces reliance on diagnostic label priors and directly probes whether models can align, localize, and complete cross-scale visual trajectories. To capture the core visual operations involved in practical pathology interpretation, we define five task types: \textit{Zoom-In Correspondence} and \textit{Zoom-Out Correspondence} identify corresponding higher- and lower-power views, respectively; \textit{Origin Grounding} localizes a crop within its parent ROI; \textit{Scale Bridging} recovers the missing intermediate view between 10$\times$ and 200$\times$; and \textit{Scale Anticipation} selects the high-power appearance most consistent with lower-power evidence. Together, these tasks complement semantic reasoning by assessing the visual alignment and trajectory-level consistency underlying cross-scale diagnosis.

\noindent\textbf{Structure-controlled Distractor Sampling.}
Although image-option questions can avoid text-only leakage, they remain vulnerable to a different form of shortcut. Random distractors may be distinguishable by staining variation, scanner-specific appearance, tissue coverage, background content, image sharpness, or crop artifacts. To address this risk, we adopt a structure-controlled distractor sampling pipeline based on the expert-verified diagnostic paths (Fig.~\ref{fig:main}B). First, we organize annotated ROIs into a path-indexed visual registry that records the diagnostic path ID per WSI, cross-scale ROI links, and parent-child relationships. For each grounding question, the anchor and correct option are drawn from the same diagnostic path, ensuring that correctness corresponds to an expert-validated trajectory. Distractors are then sampled according to the task objective. For spatial alignment tasks, including \textit{Zoom-In}, \textit{Zoom-Out Correspondence}, and \textit{Origin Grounding}, distractors are drawn from other diagnostic paths within the same WSI. This controls slide-level appearance factors while requiring the model to identify the correct spatial correspondence or parent-child mapping. Candidate distractors are further constrained to valid tissue regions based on tissue segmentation and low IoU with the ground-truth region. For trajectory completion tasks, including \textit{Scale Bridging} and \textit{Scale Anticipation}, the focus is on morphological consistency across scales. Distractors are therefore sampled from different WSIs since the objective is to select the view that best completes or follows a diagnostic trajectory.  
We also filter distractors whose image descriptions are highly compatible with the anchor, reducing ambiguity while preserving visually challenging alternatives. This design thus provides a visual-focused VQA set to evaluate whether VLMs can correctly match corresponding regions and features across scales.

\subsection{Cross-scale Semantic Reasoning Optimization}
Building on the curated \textit{cross-scale semantic reasoning} set, we develop \model{} through the two-stage optimization framework illustrated in Fig.~\ref{fig:main}(C). The first stage performs supervised fine-tuning (SFT) on difficult questions with distilled reasoning traces, providing supervision for cross-scale evidence synthesis. The second stage applies reinforcement learning (RL) to improve answer correctness while encouraging the response to follow the visual evidence structure. 

\subsubsection{SFT via Difficulty-Driven Reasoning Distillation}
\textbf{Difficulty-based Question Selection.}
Motivated by recent findings that model complex reasoning ability can emerge through small, high-quality examples acting as "cognitive templates"~\cite{limo,lima}, we adopt a difficulty-driven filtering strategy. By identifying questions with consistently low success rates across repeated inference, this strategy helps concentrate supervision on cross-scale reasoning patterns that remain challenging for the base model.
Formally, for each sample $(I,q,\mathcal{O},a^\ast)\in\mathcal{D}_{\mathrm{sem}}$, where $I$ denotes the input image set, $q$ is the question, $\mathcal{O}$ is the answer option set, and $a^\ast$ is the expert-verified answer, we query the baseline pathological VLM, Patho-R1~\cite{pathor1}, for $K$ attempts. We use the number of correct predictions as an empirical estimate of model solvability and define the hard subset as:
\begin{equation}\mathcal{D}_{\mathrm{hard}}=\left\{(I,q,\mathcal{O},a^\ast) \in \mathcal{D}_{\mathrm{sem}}\mid c_{\mathrm{base}}(I,q,\mathcal{O}) < \tau K\right\},
\end{equation}
where $c_{\mathrm{base}}(I,q,\mathcal{O})$ denotes the number of correct responses among $K$ attempts, and $\tau \in [0,1]$ is a difficulty threshold controlling the maximum empirical success rate allowed for inclusion. This criterion retains samples on which the baseline model has a low success rate, thereby concentrating supervised reasoning distillation on examples that better expose the model's current weaknesses.

\noindent\textbf{Reasoning Construction via Distillation.} 
For each retained hard question, we use a strong open-source teacher model, Qwen3.5-397B~\cite{qwen35}, to generate candidate reasoning traces. Since raw teacher outputs often contain redundant self-verification and repeated speculation, they are unsuitable for directly training a compact student VLM. Directly imitating such responses may dilute the core diagnostic logic and induce unstable reasoning. To address this, we normalize the teacher-generated traces into a compact and standardized format. The normalized response preserves the final answer and essential diagnostic logic, while removing redundancy and unsupported speculation. Based on manual inspection of retained high-quality traces, we organized the rationale into structured chain-of-thought rationales that include the task focus, image-wise visual analysis, key visual evidence, confirmed and uncertain findings, option evaluation, cross-scale conclusion, and final answer. This process yields a high-quality reasoning corpus that logically analyzes cross-scale evidence to reach diagnostically meaningful conclusions.

\noindent\textbf{SFT with Distilled Rationales.}
Using this hard subset and the normalized teacher rationales, we fine-tune the student model via the autoregressive supervised fine-tuning objective. The input consists of the cross-scale image set, question, and options, while the target is the distilled rationale followed by the final answer. This stage structurally teaches the model to analyze visual evidence across scales to reach a correct conclusion.
 
\subsubsection{RL Optimization with Scale-aware Reasoning Structure Reward}
After SFT establishes a structured cross-scale reasoning pattern, we further optimize the model with RL to improve answer correctness and reinforce reasoning across scales.

\noindent\textbf{Reward Design.} We formulate the reward as outcome-verifiable optimization under scale-aware reasoning constraints. For a generated response $o$, expert-verified answer $a^\ast$, and cross-scale image set $I=\{I_1,\ldots,I_N\}$, the reward is defined as:
\begin{equation}
R(o)=\lambda_{\mathrm{acc}} R_{\mathrm{acc}}(o,a^\ast)+\lambda_{\mathrm{srsr}} R_{\mathrm{srsr}}(o,I)+\lambda_{\mathrm{fmt}} R_{\mathrm{fmt}}(o),
\end{equation}
where $R_{\mathrm{acc}}(o,a^\ast) = \mathbbm{1}\left[\hat{a}(o)=a^\ast\right]$ ensures answer correctness, $R_{\mathrm{fmt}}(o)=\mathbbm{1}\left[o \in \mathcal{F}\right],$ ensures format compliance with \texttt{<think>} and \texttt{<answer>} tags. Although SFT provides image-wise reasoning demonstrations, the model may still collapse a multi-image input into a brief global description, without explicitly distinguishing the evidence contributed by individual images. This behavior is undesirable for cross-scale pathological reasoning, where different magnifications provide complementary morphological information. We therefore introduce a Scale-aware Reasoning Structure Reward ($R_{srsr}$) to reinforce explicit image-wise analysis before cross-scale synthesis. Let $o^{\mathrm{ana}}$ denote the analysis segment extracted from the generated response. We apply a deterministic structure extractor $\phi(\cdot)$ to identify the ordered sequence of image indices explicitly referenced in the analysis: $\phi\!\left(o^{\mathrm{ana}}\right)=(m_1,\ldots,m_L)$. Here, the sequence is determined by the first occurrence of each image reference in the generated analysis. Given the expected scale sequence
$\mathcal{S}(I)=(1,\ldots,N)$, the structure reward is defined as
\begin{equation}
R_{\mathrm{srsr}}(o,I)= 
\begin{cases}
1, & \mathcal\phi\!\left(o^{\mathrm{ana}}\right)=\mathcal{S}(I), \\
0, & \text{otherwise}.
\end{cases}
\end{equation}
\noindent This reward is assigned only when the generated reasoning accounts for all provided visual contexts in their intended scale order. As a result, $R_{\mathrm{srsr}}$ penalizes hallucinated visual references, or inconsistent cross-scale organization, thereby encouraging more complete and organized image-wise analysis.

\noindent\textbf{Policy
Optimization.} We optimize the model using standard Group Relative Policy Optimization (GRPO)~\cite{deepseekmath} to maximize the proposed reward. The final optimized model, \model{}, combines difficulty-driven reasoning distillation with scale-aware reward optimization to improve cross-scale pathological evidence integration.

\section{Experiments and Results}
\label{sec:experiments}
\subsection{Experiment Setting}
\noindent\textbf{Datasets.}
Our curated \dataset{} contains \mcq\ cross-scale VQA samples from \triplets\
expert-verified diagnostic paths extracted from \wsis\ TCGA WSIs, spanning fourteen organ sites (Fig.~\ref{fig:stats}A). The semantic reasoning subset contains 6,030 samples, of which 3,680 are used for training and validation, and 2,350 are used for testing. All splits are constructed at the patient-wise WSI level. The visual grounding subset contains 4,343 samples and is used exclusively for evaluation. We also evaluate on PathMMU~\cite{pathmmu}, a public single-scale pathology VQA benchmark with 10,387 questions, to assess transferability beyond the proposed cross-scale setting.
\begin{table}[t!]
\caption{Overall results of models on Cross-scale Semantic Reasoning. The best performance in each column is
highlighted in \textbf{bold}, the second-best is
\underline{underlined}.}
\label{tab:cross-scale-res}
\centering
\resizebox{\linewidth}{!}{%
\begin{tabular}{@{}l>{\columncolor{blue!10}}cccccc@{}}
        \toprule
        \multirow{2}{*}{\textbf{Model}}
        & \textbf{Overall}
        & \textit{\textbf{Corr.}}
        & \textit{\textbf{Conf.}}
        & \textit{\textbf{Loca.}}
        & \textit{\textbf{Expl.}}
        & \textit{\textbf{Diag.}} \\
        & \scriptsize{(2350)}
        & \scriptsize{(470)}
        & \scriptsize{(470)}
        & \scriptsize{(470)}
        & \scriptsize{(470)}
        & \scriptsize{(470)} \\        
        \midrule
        \textbf{\textit{General VLM}}
        & \multicolumn{6}{l}{} \\
        \hdashline[2pt/3pt]
        \addlinespace
        Qwen2.5-VL-7B~\cite{qwen25vl}
        & 55.96 & 47.23 & 48.30 & 77.02 & 39.57 & 67.66 \\
        MiMo-VL-7B~\cite{mimovl}
        & 47.45 & 36.17 & 43.62 & 54.47 & 38.30 & 64.68 \\
        Qwen3-VL-8B~\cite{qwen3vl}
        & 59.83 & 48.30 & 61.28 & 75.74 & 36.17 & \underline{77.66} \\
        InternVL3.5-8B~\cite{internvl35}
        & \underline{64.13} & \underline{53.19} & \underline{69.57} & 78.94 & \underline{44.47}
        & 74.47 \\
        \midrule
        \textbf{\textit{Medical VLM}}
        & \multicolumn{6}{l}{} \\
        \hdashline[2pt/3pt]
        \addlinespace
        LLaVA-Med-7B~\cite{llavamed}
        & 22.21 & 27.23	& 15.96	& 23.19	& 21.91 & 22.77 \\          
        HuatuoGPT-7B~\cite{huatuogpt}
        & 50.13 & 35.74 & 38.51 & 74.68 & 39.15 & 62.55 \\
        QoQ-Med-VL-7B~\cite{qoqmed}
        & 52.98 & 46.38 & 41.49 & 74.68 & 33.83 & 68.51 \\
        Lingshu-7B~\cite{lingshu}
        & 56.26 & 39.57 & 54.89 & 71.91 & 41.06 & 73.83 \\
        MedVLThinker-7B~\cite{medvlthinker}
        & 50.72 & 41.28 & 40.85 & 71.28 & 39.36 & 60.85 \\
        OctoMed-7B~\cite{octomed}
        & 59.62 & 36.60 & 64.04 & \underline{80.85} & 43.83 & 72.77 \\
        HealthGPT-8B~\cite{healthgpt}
        & 57.28 & 30.21 & 58.94 & 78.51 & \underline{44.47} & 74.26 \\
        \midrule
        \textbf{\textit{Pathological VLM}}
        & \multicolumn{6}{l}{} \\
        \hdashline[2pt/3pt]
        \addlinespace
        Quilt-LLaVA~\cite{quilt}
        & 38.85 & 39.79 & 32.13 & 48.51 & 38.09 & 35.74 \\
        CLOVER~\cite{clover}
        & 56.77 & 38.51 & 60.43 & 78.72 & 39.79 & 66.38 \\
        Patho-R1~\cite{pathor1}
        & 50.21 & 31.49 & 34.68 & 67.23 & 44.04 & 73.62 \\
        \rowcolor{pinky}
        \textbf{\model{}}
        & \textbf{83.32}
        & \textbf{80.85}
        & \textbf{91.91}
        & \textbf{88.72}
        & \textbf{65.74}
        & \textbf{89.36} \\
        \bottomrule
        \end{tabular}
    }
\end{table}
\begin{table*}[!]
	\caption{Overall results of models on the PathMMU. The best performance is highlighted in \textbf{bold}, the second-best is \underline{underlined}.}
    \label{tab:pathmmu}
	\centering
    \resizebox{\textwidth}{!}{%
		\begin{tabular}{@{}l>{\columncolor{blue!10}}c>{\columncolor{blue!10}}c>{\columncolor{blue!10}}ccccccccccc@{}}
    	\toprule
        \multirow{3}{*}{\textbf{Model}} 
        & \multicolumn{3}{>{\columncolor{blue!10}}c}{\textbf{Overall}} 
        & \multicolumn{2}{c}{\textbf{\textit{PubMed}}} 
        & \multicolumn{2}{c}{\textbf{\textit{SocialPath}}} 
        & \multicolumn{2}{c}{\textbf{\textit{EduContent}}} 
        & \multicolumn{2}{c}{\textbf{\textit{Atlas}}} 
        & \multicolumn{2}{c}{\textbf{\textit{PathCLS}}} \\

        \cmidrule(r{0.25em}){2-4}
        \cmidrule(l{0.25em}r{0.25em}){5-6}
        \cmidrule(l{0.25em}r{0.25em}){7-8}
        \cmidrule(l{0.25em}r{0.25em}){9-10}
        \cmidrule(l{0.25em}r{0.25em}){11-12}
        \cmidrule(l{0.25em}r{0.25em}){13-14}

        & \textbf{Val} & \textbf{Test-Tiny} & \textbf{Test} 
        & \textbf{Test-Tiny} & \textbf{Test} 
        & \textbf{Test-Tiny} & \textbf{Test} 
        & \textbf{Test-Tiny} & \textbf{Test}  
        & \textbf{Test-Tiny} & \textbf{Test}  
        & \textbf{Test-Tiny} & \textbf{Test} \\

        & \scriptsize{(710)} & \scriptsize{(1156)} & \scriptsize{(8521)}
        & \scriptsize{(281)} & \scriptsize{(2787)}
        & \scriptsize{(235)} & \scriptsize{(1620)}
        & \scriptsize{(255)} & \scriptsize{(1683)}
        & \scriptsize{(208)} & \scriptsize{(799)}
        & \scriptsize{(177)} & \scriptsize{(1632)} \\

        \midrule
		\textbf{General VLM} & \multicolumn{13}{l}{} \\ 
		\hdashline\addlinespace

		Qwen2.5-VL-7B~\cite{qwen25vl}      
        & 44.3 & 48.2 & 44.4 & 53.4 & 50.5 & 50.0 & 47.7 & 59.6 & 48.9 & 48.6 & 46.8 & 29.4 & 28.0 \\

        MiMo-VL-7B~\cite{mimovl} 
        & 36.9 & 37.6 & 35.8 & 48.8 & 40.2 & 45.4 & 41.3 & 42.0 & 41.9 & 36.1 & 40.3 & 15.8 & 15.5 \\

        Qwen3-VL-8B~\cite{qwen3vl} 
        & 49.4 & 53.9 & 52.9 & 63.7 & 56.8 & 58.3 & 54.4 & 53.7 & 54.9 & 51.0 & 56.7 & 42.9 & 41.9 \\

        InternVL3.5-8B~\cite{internvl35}     
        & 59.0 & 63.9 & 59.7 & \underline{70.5} & 63.9 & 62.0 & 65.2 & 69.4 & 61.6 & 69.2 & 67.2 & 40.8 & 40.6 \\

        \midrule
        \textbf{Medical VLM} & \multicolumn{13}{l}{} \\ 
        \hdashline\addlinespace

        LLaVA-Med-7B~\cite{llavamed}    
        & 17.5 & 22.2 & 22.7 & 24.6 & 25.1 & 23.4 & 21.1 & 25.1 & 23.6 & 17.8 & 24.5 & 20.3 & 19.2 \\

        HuatuoGPT-7B~\cite{huatuogpt}  
        & 43.2 & 43.8 & 41.2 & 49.1 & 47.7 & 52.3 & 46.0 & 52.9 & 46.1 & 44.7 & 46.9 & 19.8 & 19.2 \\

        QoQ-Med-VL-7B~\cite{qoqmed}   
        & 45.9 & 47.4 & 46.4 & 55.5 & 51.3 & 51.9 & 49.5 & 52.2 & 50.6 & 44.7 & 46.1 & 32.8 & 34.4 \\

        Lingshu-7B~\cite{lingshu}      
        & 51.9 & 56.0 & 53.7 & 60.5 & 57.6 & 61.6 & 58.1 & 69.4 & 58.7 & 58.1 & 62.5 & 30.4 & 31.6 \\

        MedVLThinker-7B~\cite{medvlthinker} 
        & 45.8 & 49.9 & 46.1 & 56.2 & 50.3 & 54.6 & 50.2 & 55.7 & 49.7 & 51.4 & 53.2 & 31.6 & 27.3 \\

        OctoMed-7B~\cite{octomed}      
        & 53.0 & 59.4 & 55.8 & 69.4 & 61.5 & 62.5 & 56.9 & 65.1 & 56.9 & 56.7 & 60.1 & \underline{43.5} & \underline{43.5} \\

        HealthGPT-8B~\cite{healthgpt} 
        & 53.8 & 63.4 & 60.0 & 69.8 & 64.6 & \underline{68.5} & 64.6 & 68.2 & 62.0 & 70.2 & 67.0 & 40.1 & 41.7 \\

        \midrule
        \textbf{Pathological VLM} & \multicolumn{13}{l}{} \\ 
        \hdashline\addlinespace

        Quilt-LLaVA~\cite{quilt} 
        & 33.8 & 32.3 & 31.9 & 30.3 & 34.2 & 26.8 & 34.6 & 35.7 & 33.3 & 41.8 & 33.3 & 27.1 & 24.0 \\

        CLOVER~\cite{clover} 
        & 53.1 & 60.6 & 56.6 & 68.0 & 59.8 & 67.6 & 60.5 & 68.6 & 62.2 & 62.0 & 64.1 & 36.7 & 36.6 \\

        Patho-R1~\cite{pathor1} 
        & \underline{62.3} 
        & \underline{64.8} 
        & \underline{62.6} 
        & 68.7 
        & \underline{64.9} 
        & 63.9 
        & \underline{65.7} 
        & \underline{71.0} 
        & \underline{66.1} 
        & \underline{78.4} 
        & \underline{73.5} 
        & 41.8 
        & 42.7 \\

        \rowcolor{pinky}{\textbf{\model}} 
        & \textbf{63.6} 
        & \textbf{70.1} 
        & \textbf{64.5} 
        & \textbf{73.0} 
        & \textbf{68.0} 
        & \textbf{73.6} 
        & \textbf{67.1} 
        & \textbf{78.0} 
        & \textbf{68.0} 
        & \textbf{80.3} 
        & \textbf{75.2} 
        & \textbf{45.8} 
        & \textbf{44.1} \\

		\bottomrule
	\end{tabular}
}
\end{table*}
\begin{table}[t]
\centering
\caption{Image-ablation analysis on the cross-scale semantic reasoning test set. Values in parentheses denote accuracy drops relative to using all images.}
\label{tab:drop_image_ablation}
\resizebox{\linewidth}{!}{%
\begin{tabular}{lcccccc}
\toprule
 & \textbf{\textit{Corresp.}} 
 & \textbf{\textit{Confirm.}} 
 & \textbf{\textit{Localiz.}} 
 & \textbf{\textit{Explana.}} 
 & \textbf{\textit{Diagnos.}} 
 & \textbf{\textit{AVG}} \\
\midrule
Full images 
& 49.83 & 69.15 & 80.55 & 46.94 & 77.70 & 64.83\\
\midrule
\multirow{2}{*}{Drop 1 image}
& 44.28 & 53.36 & 69.40 & 43.03 & 66.67 & 55.35 \\
& {\scriptsize\color{red}($\downarrow$5.55)}& {\scriptsize\color{red}($\downarrow$15.79)}
& {\scriptsize\color{red}($\downarrow$11.15)}
& {\scriptsize\color{red}($\downarrow$3.91)}
& {\scriptsize\color{red}($\downarrow$11.03)}
& {\scriptsize\color{red}($\downarrow$9.48)}\\
\midrule
\multirow{2}{*}{Drop 2 images}
& 33.42 & 46.93 & 60.29 & 38.68 & 58.28 & 47.52 \\
& {\scriptsize\color{red}($\downarrow$16.41)}& {\scriptsize\color{red}($\downarrow$22.22)}
& {\scriptsize\color{red}($\downarrow$20.26)}
& {\scriptsize\color{red}($\downarrow$8.26)}
& {\scriptsize\color{red}($\downarrow$19.42)}
& {\scriptsize\color{red}($\downarrow$17.31)}\\
\midrule
\multirow{2}{*}{\makecell[l]{Drop 3 images\\(Text-only)}}
& 17.40 & 27.86 & 34.48 & 20.89 & 38.30 & 27.79 \\
& {\scriptsize\color{red}($\downarrow$32.43)}& {\scriptsize\color{red}($\downarrow$41.29)}
& {\scriptsize\color{red}($\downarrow$46.07)}
& {\scriptsize\color{red}($\downarrow$26.05)}
& {\scriptsize\color{red}($\downarrow$39.40)}
& {\scriptsize\color{red}($\downarrow$37.04)}\\
\bottomrule
\end{tabular}
}
\end{table}

\noindent\textbf{Implementation Details.}
\model{} uses Patho-R1-7B~\cite{pathor1} as its backbone, which was trained on large-scale pathology data but focuses only on single-scale analysis. For difficulty filtering, we set $K$= 16, $\tau$= 0.25. During SFT, we freeze the vision tower and train for 5 epochs with a learning rate of 1e-4. For RL, the policy is initialized from the SFT model and optimized with GRPO for 700 steps with 16 sampled responses per prompt. The reward weights are set to $\lambda_{\mathrm{acc}}$= 0.75, $\lambda_{\mathrm{srsr}}$= 0.20, and $\lambda_{\mathrm{fmt}}$= 0.05. To provide a comprehensive and fair comparison, we compare with recent general-purpose VLMs~\cite{qwen25vl,qwen3vl,internvl35, mimovl}, medical VLMs~\cite{llavamed,huatuogpt,qoqmed,lingshu,medvlthinker,octomed,healthgpt}, and pathology-specific VLMs~\cite{quilt,clover,pathor1} with parameter scales comparable to our 7B model.  All models are evaluated using the same prompt template, image order, decoding settings, and answer-parsing procedure. Accuracy (\%) is used as the evaluation metric for all VQA tasks.

\subsection{Performance on Cross-Scale Semantic Reasoning}
We first evaluate the models on cross-scale semantic reasoning, which directly measures their ability to integrate pathological evidence across magnifications. Table~\ref{tab:cross-scale-res} reports the overall accuracy and task-wise performance across five reasoning dimensions. \model{} achieves the best overall accuracy of 83.32\%, substantially outperforming the evaluated general-purpose, medical, and pathology-specific VLMs. Compared with the strongest general-domain model, InternVL3.5-8B, {\model} improves the overall accuracy by 19.19\%. It also exceeds the strongest medical baseline, OctoMed-7B,  and the strongest pathology-specific baseline, CLOVER, by 23.70\% and 26.55\%, respectively. Relative to the Patho-R1 backbone, \model{} improves overall accuracy from 50.21\% to 83.32\%, with consistent gains across all five reasoning dimensions. The particularly large gains on \textit{Correspondence} and \textit{Confirmation} indicate an improved ability to associate observations across magnifications and determine whether evidence at one scale supports findings observed at another. The results demonstrate that difficulty-driven reasoning distillation and scale-aware RL strengthen the cross-scale evidence integration targeted by the proposed framework.
\subsection{Transfer Capability to Single-Scale VQA}
Although the proposed framework is developed for multi-image cross-scale reasoning, its learned diagnostic reasoning strategy may also benefit conventional pathology tasks involving a single image. We therefore examine \model{} on the PathMMU~\cite{pathmmu} validation and test sets, with the results reported in Table~\ref{tab:pathmmu}. \model{} achieves the best overall performance across all evaluation splits, obtaining 63.6\% accuracy on the validation set, 70.1\% on the test-tiny set, and 64.5\% on the full test set. Compared with the Patho-R1 backbone, these results correspond to improvements of 1.3\%, 5.3\%, and 1.9\%, respectively. The gains are also consistent across the different data sources in PathMMU, outperforming strong general-domain, medical-domain, and pathology-domain baselines. These results suggest that learning from diagnostic paths appears to improve how the model identifies and organizes pathological evidence, which remains useful even when only one scale is available. Thus, cross-scale reasoning supervision not only improves multi-image evidence integration but also strengthens pathology understanding on conventional single-scale VQA.

\subsection{Effectiveness of Shortcut-Resistant VQA Curation}
To verify the effectiveness of our shortcut-resistant curation, we evaluate whether the \dataset{}-Semantic questions genuinely require visual and cross-scale evidence through a progressive image removal setting. To reduce model-specific bias and avoid floor effects from weak models, we select the five strongest-performing models: \model{}, InternVL3.5-8B, Qwen3-VL-8B, OctoMed-7B, and HealthGPT-8B. For each question, we retain the original question and answer options while progressively removing views from the diagnostic path. The results are averaged across the five models under each condition. As shown in Table~\ref{tab:drop_image_ablation}, average accuracy decreases from 64.83\% with all images to 55.35\% after removing one image and 47.52\% when only one image remains. This consistent degradation indicates that the questions depend on complementary evidence across magnifications rather than isolated visual recognition. In the text-only setting, accuracy further falls to 27.79\%. These results show that textual priors alone are insufficient to reproduce full-image performance, confirming strong dependence on cross-scale visual evidence.

\subsection{Ablation Analysis of the Key Components}
We conduct ablations to examine the contributions of difficulty-driven SFT and the scale-aware reasoning structure reward in RL optimization. All ablation experiments are conducted on the same backbone, Patho-R1-7B, to ensure a controlled comparison. The results are presented in Table~\ref{tab:ablation_reward}.

\noindent\textbf{\textit{Effectiveness of Difficulty-driven SFT.}} 
Using distilled rationales from difficult samples substantially improves cross-scale semantic reasoning, increasing the average accuracy from 50.21\% to 71.36\%. The largest gains occur in \textit{Correspondence} and \textit{Confirmation}, which improve from 31.49\% to 58.51\% and from 34.68\% to 83.62\%, respectively. This indicates that targeted supervision on repeatedly failed cases efficiently addresses the backbone's weaknesses in associating and verifying evidence across scales. 
The distilled rationales provide compact cognitive templates for associating pathological evidence across magnifications, enabling the model to better follow structured diagnostic logic. 

\noindent\textbf{\textit{Effectiveness of RL Optimization.}} Applying RL after SFT further improves the average accuracy from 71.36\% to 80.97\%. The largest gains are observed in \textit{Correspondence} and
\textit{Explanation}, with improvements of 21.06\% and 12.55\%, respectively. This suggests that SFT establishes the initial cross-scale reasoning pattern, whereas subsequent outcome-based optimization improves the model's ability to apply this pattern consistently and produce diagnostically correct conclusions. This supports our two-stage design: SFT first teaches the model how to reason over cross-scale evidence, while RL further aligns the generated reasoning and final answer with the desired diagnostic outcome. 

\noindent\textbf{\textit{Effectiveness of the Scale-aware Reasoning Structure Reward.}} Introducing $R_{\mathrm{srsr}}$ further increases the average accuracy from 80.97\% to 83.32\%. Improvements are observed across all task categories, ranging from 1.28\% in \textit{Correspondence} to 2.78\% in \textit{Confirmation}, with gains of 2.76, 2.34, and 2.55\% in \textit{Localization}, \textit{Explanation}, and \textit{Diagnosis}, respectively. The broadly distributed gains indicate that the reward provides a general benefit across different cross-scale tasks. This improvement indicates that explicitly rewarding the use of cross-scale evidence complements answer-level supervision and promotes more reliable evidence organization. 

\begin{figure*}[t!]
    \centering
    \includegraphics[width=1\textwidth]{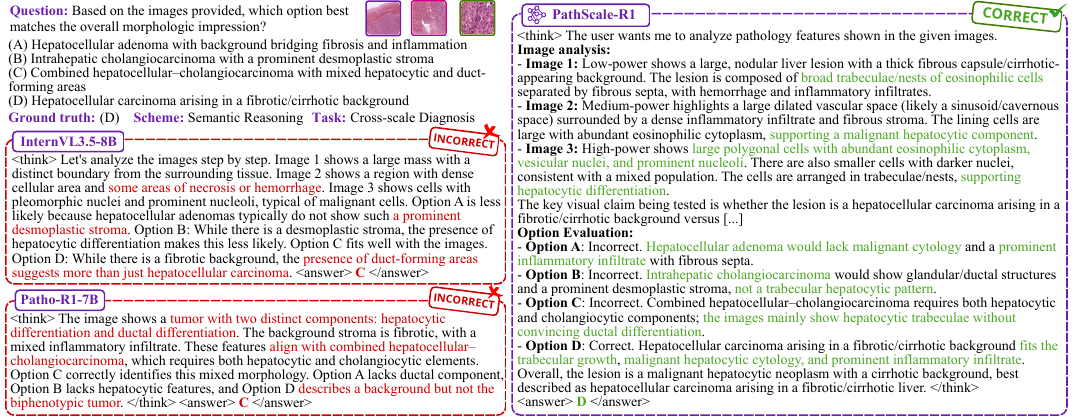}
    \caption{\textbf{Cross-scale semantic reasoning VQA case study.} \model{} integrates architectural, stromal, and cellular evidence across magnifications and reaches the correct diagnosis, whereas the baseline models infer unsupported features and produce incorrect predictions.} 
    \label{fig:case-semantic}
\end{figure*} 
\begin{figure*}[t!]
    \centering
\includegraphics[width=1\linewidth]{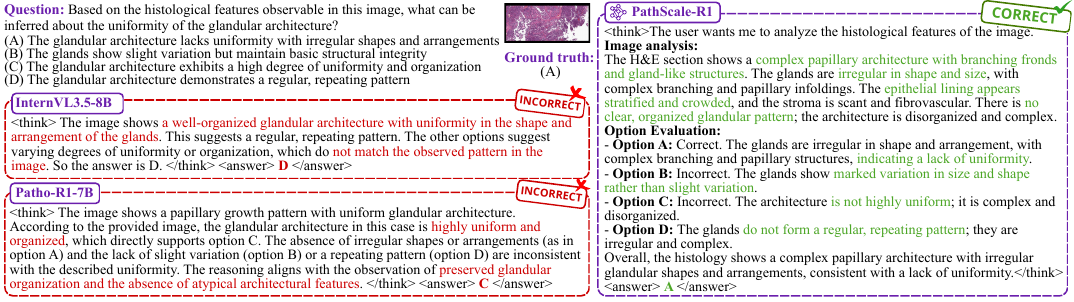}
    \caption{\textbf{Single-scale pathology VQA case study.} \model{} identifies the key morphological evidence and selects the correct answer, whereas baseline models rely on coarse visual impressions and produce incorrect answers.}  
    \label{fig:case-single}
\end{figure*} 
\subsection{Qualitative Comparison}
To qualitatively evaluate the reasoning behavior of our approach, we present representative cases comparing \model{} with the second-best-performing model, InternVL3.5-8B, and its Patho-R1 backbone. As shown in Fig.~\ref{fig:case-semantic}, the cross-scale diagnosis case requires weighing evidence from all three magnifications before reaching the final interpretation. Both baselines produce fluent pathological descriptions, yet each fixes on a locally salient impression and concludes a combined hepatocellular–cholangiocarcinoma without verifying whether ductal differentiation is actually supported at high power. \model{} instead proceeds through the images in low-to-high magnification order, records what each scale contributes, and explicitly notes the absence of convincing ductal differentiation before rejecting that option. This example illustrates that our model produces a more coherent cross-scale interpretation by relating evidence across views rather than relying on isolated findings or diagnostic priors. Fig.~\ref{fig:case-single} further shows that this structured reasoning behavior transfers to single-scale VQA. \model{} correctly identifies the irregular and complex glandular architecture and provides an option-level justification grounded in the observed morphology, whereas the baselines rely on coarse visual impressions. Taken together, these cases support the quantitative results, demonstrating stronger cross-scale evidence integration and more reliable morphology-based reasoning in both cross-scale and single-scale settings.
\begin{table}[t]
\centering
\caption{Component ablation of difficulty-driven SFT and RL optimization. The best performance in each column is \textbf{bold}.}
\label{tab:ablation_reward}
\resizebox{\linewidth}{!}{%
\begin{tabular}{@{}lcccccc@{}}
\toprule
& \textbf{\textit{Corresp.}}
& \textbf{\textit{Confirm.}}
& \textbf{\textit{Localiz.}}
& \textbf{\textit{Explana.}}
& \textbf{\textit{Diagnos.}}
& \textbf{\textit{AVG}}\\
\midrule

Baseline
& 31.49 & 34.68 & 67.23 & 44.04 & 73.62 & 50.21 \\

+ SFT
& 58.51 & 83.62	& 81.70 & 50.85 & 82.13 & 71.36 \\
+ SFT + RL (w/o $R_{\mathrm{srsr}}$)
& 79.57	& 89.13	& 85.96	& 63.40	& 86.81	& 80.97 \\

+ SFT + RL (w/ $R_{\mathrm{srsr}}$)
& \textbf{80.85}
& \textbf{91.91}
& \textbf{88.72}
& \textbf{65.74}
& \textbf{89.36} 
& \textbf{83.32} 
\\

\bottomrule
\end{tabular}
}
\end{table}
\subsection{Cross-Scale Visual Grounding: Performance and Remaining Challenges}
\begin{figure*}[!]
    \centering
    \includegraphics[width=1\linewidth]{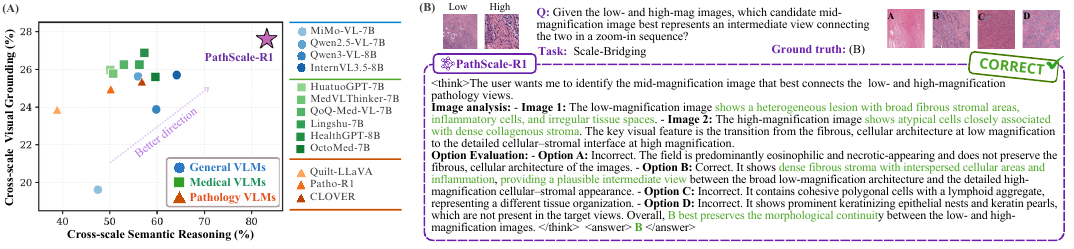}
    \caption{\textbf{Cross-scale visual grounding analysis.} (A) Comparison of cross-scale semantic reasoning and visual grounding performance across models. \model{} achieves the strongest semantic reasoning performance and the highest grounding accuracy, while grounding performance remains limited across all evaluated models. (B) A representative case in which \model{} identifies the intermediate-magnification view by preserving morphological continuity between the low- and high-magnification images.}
    \label{fig:visual-grounding}
\end{figure*}

We further evaluate models on the evaluation-only cross-scale visual grounding, which tests whether a model can identify spatial and morphological correspondence across diagnostic trajectories. As shown in Fig.~\ref{fig:visual-grounding}(A), \model{} achieves the highest overall grounding performance among the evaluated models and improves over its Patho-R1 backbone without any grounding-specific training. Figure~\ref{fig:visual-grounding}(B) illustrates a successful case study, in which \model{} identifies the intermediate view by matching stromal organization and cellular morphology across magnifications. This suggests that semantic cross-scale optimization provides a certain degree of transfer to visual trajectory understanding. Nevertheless, grounding performance remains concentrated within a relatively low and narrow range across all general, medical, and pathology-specific VLMs, despite substantially larger differences in semantic reasoning accuracy. Cross-scale visual grounding can thus be positioned as a challenging and clinically relevant capability gap that should be further addressed in future pathology VLM development.

\subsection {Limitations and Future Work}
Despite the strong cross-scale reasoning performance, several directions remain for further development. First, \dataset{} is constructed from TCGA H\&E slides and predefined 10$\times$, 40$\times$, and 200$\times$ trajectories; extending it to multi-center data and more flexible magnification paths would further broaden its coverage. Second, the current MCQ formulation enables controlled evaluation but does not fully represent open-ended diagnosis or interactive WSI navigation. Moreover, our optimization focuses primarily on semantic reasoning, while the visual grounding results indicate that fine-grained spatial correspondence remains an important capability to improve. Future work will therefore explore joint semantic-grounding optimization and validation in broader, clinically realistic settings.
\section{Conclusion}
In this work, we introduced a cross-scale pathology VQA paradigm that better aligns with the multi-magnification workflow of pathological diagnosis. We constructed high-quality \dataset{} from pathologist-verified diagnostic paths and designed complementary semantic reasoning and visual grounding tasks with shortcut-resistant curation. We further developed \model{}, combining difficulty-driven reasoning distillation with scale-aware RL optimization. Experiments show that \model{} substantially improves cross-scale semantic reasoning and transfers effectively to single-scale pathology VQA, demonstrating the practical value of cross-scale supervision. At the same time, our visual grounding benchmark reveals that fine-grained cross-scale correspondence remains challenging for current VLMs. Overall, our study highlights cross-scale reasoning as an important direction for clinically aligned pathology VLMs and provides a comprehensive benchmark and training framework for advancing future pathological AI systems.

\bibliographystyle{IEEEtran}
\bibliography{references}

\end{document}